\documentclass[letterpaper]{article} 
\usepackage[]{aaai23}  
\usepackage{times}  
\usepackage{helvet}  
\usepackage{courier}  
\usepackage[hyphens]{url}  
\usepackage{graphicx} 
\usepackage{natbib}  
\usepackage{amsmath}
\usepackage{caption} 
\usepackage{algorithm} 
\usepackage{algorithmic}  
\usepackage[algo2e]{algorithm2e}
\SetKwInput{KwInput}{Input}                
\SetKwInput{KwOutput}{Output}  

\usepackage{newfloat}
\usepackage{listings}
\DeclareCaptionStyle{ruled}{labelfont=normalfont,labelsep=colon,strut=off} 
\floatstyle{ruled}
\newfloat{listing}{tb}{lst}{}
\floatname{listing}{Listing}
\title{FedSpectral+: Spectral Clustering using Federated Learning}
\author {
    Janvi Thakkar*,
    Devvrat Joshi* 
}
\affiliations {
    IIT Gandhinagar\\
    \{janvi.thakkar, devvrat.joshi\}@iitgn.ac.in
}

\usepackage{bibentry}

\begin{document}

\maketitle
\begin{abstract}
Clustering in graphs has been a well-known research problem, particularly because most Internet and social network data is in the form of graphs. Organizations widely use spectral clustering algorithms to find clustering in graph datasets. However, applying spectral clustering to a large dataset is challenging due to computational overhead. While the distributed spectral clustering algorithm exists, they face the problem of data privacy and increased communication costs between the clients. Thus, in this paper, we propose a spectral clustering algorithm using federated learning (FL) to overcome these issues. FL is a privacy-protecting algorithm that accumulates model parameters from each local learner rather than collecting users' raw data, thus providing both scalability and data privacy. We developed two approaches: \textit{FedSpectral} and \textit{FedSpectral+}. \textit{FedSpectral} is a baseline approach that uses local spectral clustering labels to aggregate the global spectral clustering by creating a similarity graph. \textit{FedSpectral+}, a state-of-the-art approach, uses the power iteration method to learn the global spectral embedding by incorporating the entire graph data without access to the raw information distributed among the clients. We further designed our own similarity metric to check the clustering quality of the distributed approach to that of the original/non-FL clustering. The proposed approach \textit{FedSpectral+} obtained a similarity of 98.85\% and 99.8\%, comparable to that of global clustering on the \textit{ego-Facebook} and \textit{email-Eu-core} dataset. 
\end{abstract}
\section{Introduction}
Organizations worldwide actively use graph clustering algorithms to map data into communities to infer their relationships. Particularly, spectral clustering \cite{ng2001spectral} \cite{von2007tutorial} is one of the widely used algorithms to cluster graphs. It employs graph eigenvector embedding by preserving node relationships in euclidean space. However, applying spectral clustering on a graph with billions of nodes is challenging as it requires lots of computational cost and storage space. Researchers in the past had tried to distribute the graph and train the model in a distributed manner by achieving comparable accuracy. Nevertheless, in this case, the communication cost has increased exponentially \cite{macey1998performance}. In addition, \cite{chen2010parallel} also developed parallel spectral clustering based on the sparse similarity matrix. However, the major drawback of this work is that it assumes that any data in the parallel system can be accessed by any client. But, this raises concerns for data privacy if we share the information with different clients. In this work, we remove this assumption by only sharing the representations learned in the local client to the global server instead of directly communicating with neighboring clients.
\footnote{* Authors contributed equally} 
Data privacy in distributed settings was the very reason for the introduction of federated learning. The new machine learning algorithms were becoming complex and required a large amount of data to generalize on a task. However, in reality, data is distributed over many organizations, and sharing them is difficult as it concerns data privacy. Thus, the ‘Federated Learning (FL)’ term coined by Google \cite{mcmahan2017communication} was the first to propose an algorithm that combines intelligence across the clients without compromising data security. There are two key advantages of using FL over traditional machine learning algorithms. First, it allows training the models on local devices and then sending the trained parameters to the central server; this provides privacy to local learners as there is no exchange of raw data among clients. Secondly, models are trained over different local learners and thus do not need a large dataset to be present in the central cloud. 

Researchers around the globe have tried to propose several federated learning algorithms for different machine learning models. They have tried to integrate the differential privacy-based approaches in the distributed setting. Some of the past literature works includes logistic regression \cite{chen2018privacy} \cite{nikolaenko2013privacy}, deep neural networks \cite{mcmahan2017communication} \cite{yang2019federated} \cite{bonawitz2019towards}, support vector machines \cite{smith2017federated} and gradient boosted decision trees \cite{cheng2021secureboost} \cite{li2020practical}.

Thus, we propose \textit{FedSpectral+}, a novel algorithm for federated spectral clustering. The approach integrates the power iteration method \cite{booth2006power} and federated learning to enhance spectral clustering by decreasing the communication cost between clients and providing data privacy. Our algorithm takes advantage of the FL technique by learning the representations locally and aggregating them over the server. Firstly, the server creates a single matrix and sends it to all the clients parallelly. The clients run power iterations over this matrix using their own data and send them back to the server. This is repeated for a fixed number of rounds till the matrix converges to the global graph embedding. The proposed approach was tested on the \textit{ego-facebook} and \textit{email-Eu-core} datasets. We obtained a similarity of 98.85\% and 99.8\%, comparable to that of global clustering.



Our main contribution includes:
\begin{enumerate}
    \item We propose the state-of-the-art FL framework for spectral clustering, namely, \textit{FedSpectral+}. The approach uses the power iteration method to iteratively learn the eigenvector embedding while maintaining data privacy.
    \item We experimentally prove the effectiveness of our algorithm on \textit{ego-Facebook} and \textit{email-Eu-core} datasets and give a detailed study of factors influencing the results of \textit{FedSpectral+}.
\end{enumerate}


\section{Related Work}
The concept of federated learning(FL) has aroused a plethora of interdisciplinary studies, primarily due to its broad applicability in privacy-constrained scenarios \cite{li2020federated}. Federated Averaging (FedAvg) algorithm \cite{mcmahan2017communication}, a simple baseline, which aggregates the local client updates to obtain the global model without exchanging the raw data. \cite{lalitha2019peer} proposed employing peer-to-peer FL for graph-structured data. Further, for the link prediction task, \cite{chen2021fede} proposed FedE to exploit centralized aggregation to utilize FL over knowledge graphs. However, they lack in training the multiple knowledge graphs together. Moreover, several works have also explored the FL on graph neural networks - \cite{wang2020graphfl}, \cite{zhang2021subgraph}. But, the use of FL in spectral clustering still remains unexplored. The nearest work includes that of \cite{chen2010parallel} that proposes parallel spectral clustering, which revolves around the idea of using the sparse similarity matrix. However, the major drawback of this work is that it assumes that any data in the parallel system can be accessed from any node. But, if data is shared among nodes, this compromises individual information, raising concerns about data privacy.

Thus, in this work, we propose \textit{FedSpectral+}, a federated spectral clustering approach that globally clusters the data shared across different clients without sharing raw data. This approach handles the problem of data privacy by using the advantage of the FL technique, which only aggregates the representations learned by the clients and, thus, no direct exchange of raw data between the clients and server. In addition, these federated spectral settings also allow one to use large datasets without worrying about the computational cost, as the data is distributed over several clients.
\section{Problem Statement and Real World}
This section explains the rationale for recommending the FedSpectral+ strategy and provides a concrete example of how it can be applied in practice.

Let there be different social network platforms. Each platform has information about the unique phone number or email address of every person registered in the network, which is the primary key in their database. For every person(represented by a node), there will be a relationship between pair of nodes(represented by an edge) within the master set of nodes. Here, the master set of nodes refers to all the phone numbers/unique ids available worldwide. If a number is not registered on a particular platform, the social network platform still takes it into consideration by making zero edges of that id with other people/ids. Furthermore, we can consider edges to hold the weight equivalent to the amount of interaction between two nodes (number of messages exchanged). There is a possibility that a person interacting with another person on a particular social network might not interact with the same person on another. If a person interacts with a large number of people and has more than average information exchange, capturing that person's attention to a product can help reduce advertisement costs. It is because if that person likes a product, he/she will do the free advertisement to his/her friends. Each social network platform privately stores the interaction information due to privacy concerns. However, without the knowledge of the entire graph, it will be hard to learn the amount of interaction among individuals over the global graph. Thus, our algorithm solves this problem using the federated learning approach.

\section{Proposed Approach}
In this paper, we eliminate the drawback of privacy concerns in distributed spectral clustering by generating the eigenvector embedding of the global graph without revealing raw data to the server. Our first approach is the baseline - \textit{FedSpectral} algorithm. Here, the clients create the spectral clustering labels of the local graph data. The server receives these clustering labels, which further creates the similarity graph by aggregating the clusterings. While there is no exchange of raw information between any clients, the algorithm doesn't consider the relation of particular pair of nodes within two different clients. Due to this, the output clustering significantly deviates from the global clustering. To overcome this, we proposed the \textit{FedSpectral+} approach that uses the power iteration method to generate the client's eigenvector embeddings, which also considers the information from all other local clients by utilizing the aggregated result of the server.
This global aggregation helps us to learn the overall graph structure. Thus this approach preserves privacy and produces significantly better clustering than the baseline.


\subsection{Preliminaries}
\begin{enumerate}
    \item \textbf{global clustering:} It is the spectral clustering of the entire graph.
    \item \textbf{global aggregated clustering:} It is the clustering resulting from the federated server-side algorithms described in this section.
    \item \textbf{Power iteration:} A method to iteratively find the approximate eigen embedding of a matrix.
    \item \textbf{Similarity Graph:} The similarity graph is constructed by assigning a value to each pair of nodes by using equation \ref{eq:1}.
    \item \textbf{Overlap of graph:} The overlap of the graph is the fraction of times the same edge is repeated over the clients.
    \item \textbf{global embedding:} Actual eigenvector embedding constructed using spectral clustering on the entire graph.
\end{enumerate}

\subsection{Setting}
Here, we discuss how the graph is distributed among clients for experimental purposes. We create a null adjacency matrix for each client with the size of the total number of nodes in the graph. Then each edge is randomly assigned to the client uniformly. To regulate the overlap of a graph, i.e., the number of clients that will hold a particular edge, we assign the edge to a fixed number of random clients. Thus, creating our dataset for experimental purposes.

\begin{algorithm}[]

\DontPrintSemicolon
  
\KwInput{
    $\textbf{client}$: client identifier\\
    $\textbf{numOfClusters}$: Number of clusters to be formed in clustering the graph data\\
}
\KwOutput{\textbf{S:} List of labels of N nodes after clustering}
Internal Variables inside client's memory:\\
$\textbf{numOfNodes}$: Number of nodes in graph \\
$\textbf{L}$: Normalized Laplacian matrix of size $N^{2}$ where $N$ is number of nodes in graph \\
$eigenVectors$ = $computeEigenvectors(L)$\\
Sort eigenVectors based on eigenvalues\\
$vectorEmbedding$ = Transpose of $eigenVectors$ matrix\\
$labels$ = $KMeansClustering(vectorEmbedding)$\\
\textbf{Return:} $labels$

\caption{FedSpectral, Client Code \\ \textbf{function name: getClientLabels}}
\label{alg:alg1}
\end{algorithm}
\begin{algorithm}[]
\DontPrintSemicolon
  
  \KwInput{
  $\textbf{clientList}$: client list\\
  $\textbf{numOfClients}$: Number of clients\\
  $\textbf{numOfNodes}$: Number of nodes in graph\\
  $\textbf{numOfClusters}$: Number of clusters\\
  }
  \KwOutput{\textbf{S:} List of labels of N nodes after global clustering}
  
$\textbf{similarityGraph}$: initialize a similarity matrix of size $N^{2}$ with zeros as the similarity matrix of graph\\
\For{client in clientList}{
    labels = $getClientLabels(client,numOfClusters)$\\
    \For{i in range(N)}{
        \For{j in range(N)}{
            \textbf{if} $labels[i]==labels[j]$\\
            \hspace{0.5 cm}$similarityGraph[i][j]$\\ $+=$ $1/numOfClients$            
        }
    }
}
$L$ = $createLaplacian(similarityGraph)$\\
$eigenVectors$ = $computeEigenvectors(L)$
\\
Sort eigenVectors based on eigenvalues\\
$vectorEmbedding$ = Transpose of $eigenVectors$ matrix\\
$globalLabels$ = $KMeansClustering(vectorEmbedding)$\\
\textbf{Return:} $globalLabels$

\caption{FedSpectral, Server Code}
\label{alg:alg2}
\end{algorithm}
\subsection{\textit{FedSpectral} Algorithm (Baseline)}
Spectral clustering in distributed settings faces the problem of data privacy because it requires sharing the edge data of graphs to create the embedding. As a result, it reveals the relationship between nodes in the graph of a particular client. We propose our baseline - \textit{FedSpectral} approach, which does not share the graph's edge information between the clients or the server.\\

In this approach, each client calculates the spectral clustering of its local graph. The labels of all nodes are then delivered to the server by each client. The server then creates a similarity graph using the clustering information received from all clients. The similarity graph is constructed by assigning a value to each pair of nodes using equation \ref{eq:1}. Finally, the server performs spectral clustering on the resulting similarity graph to calculate global clustering.\\

\textit{Algorithm \ref{alg:alg1}} provides pseudo-code for the client side of this approach. It calculates the cluster labels of the graph data that is local to the client using spectral clustering. \textit{Algorithm \ref{alg:alg2}} provides pseudo-code for the server side of this approach. It gets the local spectral clustering labels from each client in the list \textit{clientList}. It then uses the labels received from the client to create a similarity graph using the rule : 
\begin{multline}similarityGraph[node_i][node_j] =\\\frac{\#\:of\:clients\:implying\:nodes (i,j)\:have\:same\: labels}{\#\:of\:clients}\label{eq:1}\end{multline} 

Finally, the server applies spectral clustering on this $similarityGraph$ and returns the global aggregated clustering.

\subsection{\textit{FedSpectral+} Algorithm}
Spectral clustering needs the entire graph to be kept in a single memory, which limits the method's use. However, sharing raw graph data with the global server and other clients raises privacy problems. The baseline - \textit{FedSpectral} technique ignores the primary idea of spectral clustering that an embedding should be constructed using the knowledge of the entire graph rather than sections of graphs within individual clients. This is because a particular node in two separate client datasets can have a different vector embedding. As a result, we are unable to employ the method of calculating spectral embedding individually for each client. We need to find a way to aggregate the eigenvector embedding of clients to learn the representation of the entire graph. In addition, we also need the approach to achieve the former without sharing information about the presence of an edge within the data of a local client. Therefore, in this approach, we use the power iteration method to iteratively learn the eigenvector embedding of the global graph while preventing the exchange of edge data amongst the clients and server.
\begin{algorithm}[]

\DontPrintSemicolon
  
  \KwInput{
    $\textbf{client}$: client identifier\\
    $\textbf{iters}$: Number of power iterations\\
    $\textbf{eigenVectors}$: A list of smallest $K$ approximate eigenvectors of global graph, dimension: $N\times K$\\
  }
  \KwOutput{$\textbf{eigenVectors:}$ Modified list of smallest $K$ approximate eigenvectors of global graph, dimension: $N \times K$}
  Internal Variables inside client's memory:\\
  $\textbf{numOfNodes}$: Number of nodes in graph \\
  $\textbf{L}$: Normalized Laplacian matrix of size $numOfNodes^{2}$ of client\\
  $\textbf{I}$: Identity matrix of size $numOfNodes^{2}$\\
  $L\:=\:I\:-\:L$\\
  \For{iter in range($iters$)}{
        $eigenVectors$ = $L\times eigenVectors$
    }
\textbf{Return:} $eigenVectors$

\caption{FedSpectral+, Client Code \\\textbf{function name: getPowerIterationClient}}
\label{alg:alg3}
\end{algorithm}
In this approach, we want to learn the embedding as close as possible to the global embedding. The clients initially learn the local embedding using the power iteration method over their Laplacian matrix and send it to the server. Now, the server aggregates(averages) all the embeddings from the local clients and calls this the approximate embedding. The server then sends this approximate embedding to the clients. The client repeats the power iteration method using this learned approximate embedding to incorporate the local graph structure. Here, each node relates to the neighboring nodes of the other clients using approximate embedding. Thus, it gives the impression that we are learning the embedding by incorporating the features of the entire graph. As the number of aggregation rounds increases, the approximate embedding becomes nearer to the global embedding since the global graph structure gets clearer at every step. In the proposed approach, the only exchange of information includes sending the approximate embedding between the server and the clients. Therefore, there is no direct sharing of raw information about the local graph data, maintaining privacy in our federated approach.\\
\begin{algorithm}[]
\DontPrintSemicolon
  
  \KwInput{
  $\textbf{clientList}$: client list\\
  $\textbf{numOfClients}$: Number of clients\\
  $\textbf{numOfNodes}$: Number of nodes in graph\\
  $\textbf{numOfClusters}$: Number of clusters\\
  $\textbf{iters}$: Number of power iteration per client\\
  $\textbf{globalRounds}$: Number of global aggregation rounds\\
  }
  \KwOutput{\textbf{S:} List of labels of N nodes after global clustering}
  
$\textbf{v}$: initialize a $numOfNodes\times numOfCluster$ matrix randomly\\
\For{round in range($globalRounds$)}{
    $\textbf{p}$ = $numOfClients$ copies of $v$\\
    \For{client in range(numOfClients)}{
        $clientId$ = $clientList[client]$\\
        $p[client]$ = $getPowerIterationClient($\\
        \hspace{1 cm}$clientId,numOfClusters,iters,p[client])$\\
     }
     $v$ = average of all matrices in $p$\\
     $v,r$ = $qrDecomposition(v)$
}

$globalLabels$ = $KMeansClustering(v)$\\
\textbf{Return:} $globalLabels$
\\
\caption{FedSpectral+, Server Code}
\label{alg:alg4}
\end{algorithm} 
Algorithmically, we randomly initialize a matrix $v$ of size $N \times K$. Given the nature of spectral clustering, if there are $K$ clusters, then the first $K$ eigenvectors of the graph's Laplacian matrix should be used to generate the nodes' vector embeddings. Therefore, after completion of the algorithm, each of the columns of $v$ will represent one of the first $K$ eigenvectors of the global graph. Let $C$ be the number of clients contributing to the clustering. We create $C$ copies of $v$ and call it $p$. Then we pass these matrices to their corresponding clients. Let  $p[i]$ be the $v$ matrix of client $i$. Clients will now run the power iteration approach to calculate the first $K$ eigenvectors simultaneously on their local graphs. After completing the power iterations, all clients will send their $p[i]$ matrix to the global server. The server will compute the average of all the $p[i]$ matrices of clients and then orthogonalize the resulting matrix using a QR decomposition. It will assign the $q$ of the QR decomposition to $v$. Again the server will share $C$ copies of this aggregated matrix $v$ to clients, and then the same process will be repeated for a number of $globalRounds$. After completing all the global rounds, we get the approximate first $K$ eigenvectors in the form of the $N \times K$ matrix, with each column representing one eigenvector. Since $v$ matrix is the eigenvector embedding of the global graph, we directly apply $KMeans$ clustering algorithm over this vector embedding and return the $globalLabels$.\\

\textit{Algorithm \ref{alg:alg3}} provides the pseudo-code for the client side of this approach. Since the power iteration method calculates the eigenvectors corresponding to the largest eigenvalues, each client applies this function to the normalized local Laplacian matrix: $L\:=\:I\:-\:L$. It then runs the power iterations over $eigenVectors$ with $L$ as the matrix multiplier. Finally, it returns the modified $eigenVectors$. For the pseudo-code of server-side of this algorithm, refer \textit{Algorithm \ref{alg:alg4}}. The QR decomposition used in this approach is taken from the NumPy library of python (\textit{numpy.linalg.qr()}).

\begin{algorithm}[]
\DontPrintSemicolon
  
  \KwInput{${\textbf{numOfNodes}}$: number of nodes in graph\\
  ${\textbf{aggregatedLabels}}$: labels from the federated algorithm\\
  ${\textbf{globalLabels}}$: labels from non distributed setting\\
  }
  \KwOutput{$clusterSimilarity$}
  $misMatch$ = 0\\
  \For{i in range(N)}{
        \For{j in range(N)}{
            \textbf{if} $globalLabels[i]==globalLabels[j]$\\
            \hspace{0.5 cm}\textbf{if} $aggregatedLabels[i]!=aggregatedLabels[j]$\\
            \hspace{0.5 cm}$misMatch$  $+=$ $1$
        }
    }
    $clusterSimilarity$ = $1-\frac{misMatch}{numOfNodes^{2}}$\\
    \textbf{$Return:$} $clusterSimilarity$
\caption{clusterSimilarityMetric}
\label{alg:alg5}
\end{algorithm}
\begin{figure*}[t]
\includegraphics[width=\linewidth, height=2in]{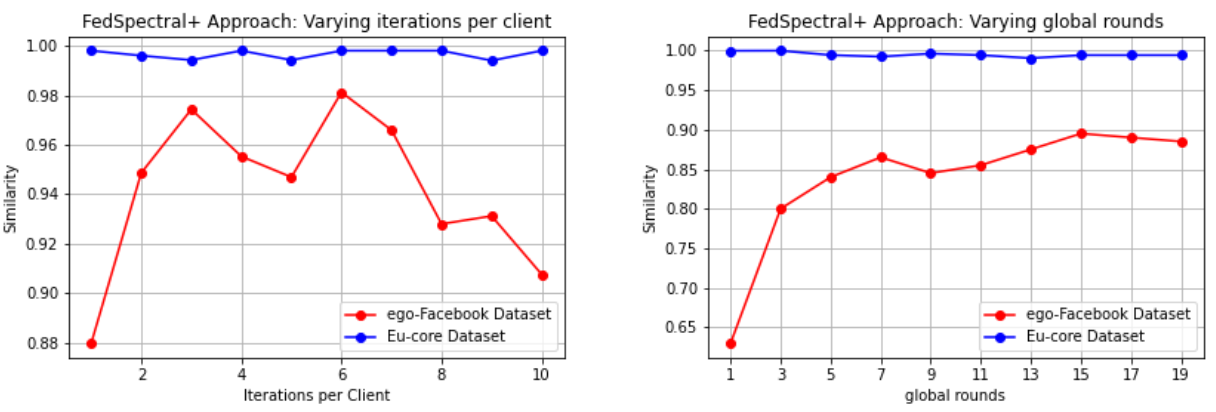}
\caption{The figure a) on the left, shows the plot of clustering similarity Vs iteration per client on two datasets. The figure b) on the right, shows the plot of clustering similarity Vs global rounds on two datasets.}
\label{fig:1.jpg}
\end{figure*}
\section{Evaluation}
\subsection{Metric}
\begin{figure*}[t]
\includegraphics[width=\linewidth, height=2in]{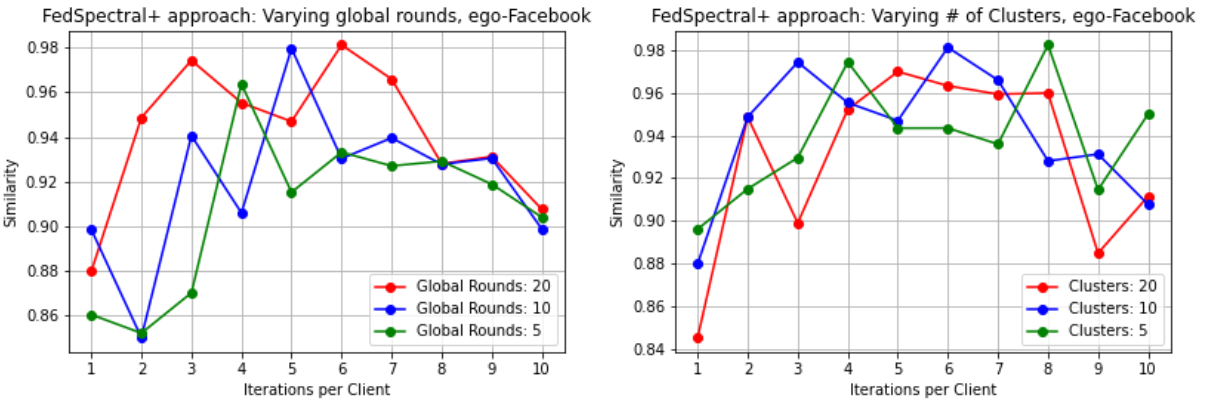}
\caption{The figure a) on the left, shows the plot of how the clustering similarity Vs iteration per client changes with respect to global rounds on \textit{ego-Facebook} dataset. The figure b) on the right, shows the plot of how the clustering similarity Vs iteration per client changes with respect to the number of clusters on \textit{ego-Facebook} dataset.}
\label{fig:2.jpg}
\end{figure*}

We have created our own metric for comparing the similarity between two different clustering of a graph. Our metric algorithm penalizes the similarity score for each pair of nodes if the pair lie in the same cluster in global clustering and lie in different clusters in global aggregated clustering. The similarity score ranges from 0 to 1. A higher similarity score represents that both the clustering are more similar. Refer to algorithm \ref{alg:alg5} for the pseudo-code of similarity score. 
\textbf{Please Note}: \textit{The y-axis in the plots \ref{fig:1.jpg}, \ref{fig:2.jpg} and \ref{fig:3.jpg} is the similarity metric defined above.}
\subsection{Datasets}
\begin{enumerate}
    \item \textbf{{ego-Facebook \cite{snapnets}}}: It is an undirected social network graph dataset consisting of 4039 nodes and 88234 edges. The data is made up of anonymized Facebook social circles.
    \item \textbf{{email-Eu-core  \cite{snapnets}}}: It is a directed network graph dataset consisting of 1005 nodes and 25571 edges. It is an email communication link dataset among the members of an organization. We converted this graph to undirected for experimental purpose.
\end{enumerate}

\subsection{Experimental Setting}
In \textit{FedSpectral} approach:
\begin{itemize}
    \item We used 5 clients and 10 clusters as the parameters.
\end{itemize}
In \textit{FedSpectral+} approach:
\begin{itemize}
    \item For both dataset, we used 5 clients and 10 clusters as the parameters.
    \item For the \textit{ego-Facebook} dataset, we set $iters$ as 6 and $globalRounds$ as 20 for the best results.
    \item For the \textit{email-Eu-core} dataset, we set $iters$ as 1 and $globalRounds$ as 1 for the best results.
\end{itemize}

\subsection{Results and Analysis}
We experimented with our \textit{FedSpectral} and \textit{FedSpectral+} approach on \textit{ego-Facebook} and \textit{email-Eu-core} datasets. 
\begin{itemize}
    \item The \textit{FedSpectral} approach achieved a clustering similarity of 77.63\% on \textit{ego-Facebook} dataset and 87.05\% on \textit{email-Eu-core} dataset respectively.
    \item The \textit{FedSpectral+} approach achieved a clustering similarity of 98.85\% on \textit{ego-Facebook} and 99.8\% on \textit{email-Eu-core} dataset respectively.
\end{itemize}

In Figure \ref{fig:1.jpg}, a) shows the variation in similarity with the varying number of iterations for \textit{ego-Facebook} and \textit{email-Eu-core} dataset with constant 20 and 1 global rounds respectively on the \textit{FedSpectral+} algorithm with graph distributed using 40\% overlap.
\begin{itemize}
    \item We can observe from the plot that as the number of iterations increases for \textit{ego-Facebook}, the similarity increases and then again drops. The increase in similarity is because the number of iterations required to converge the eigenvectors is not sufficient. However, the similarity drops because aggregation of eigenvectors is done after a larger number of power iterations. Thus, it focuses more on the local graph embedding instead of the global graph.
    \item We can observe from the plot that as the number of iterations increases for \textit{email-Eu-core} dataset, the similarity remains almost constant. Here, the number of edges in the graph is very high in comparison to \textit{ego-Facebook} the number of nodes. However, it is a small dataset, so it takes lesser iterations to converge. Since, for each client, the local graph requires only one iteration to converge to the local embedding, increasing the iterations does not affect the similarity as the number of global rounds is fixed to 1.
\end{itemize}
\begin{figure*}[t]
\includegraphics[width=\linewidth, height=2in]{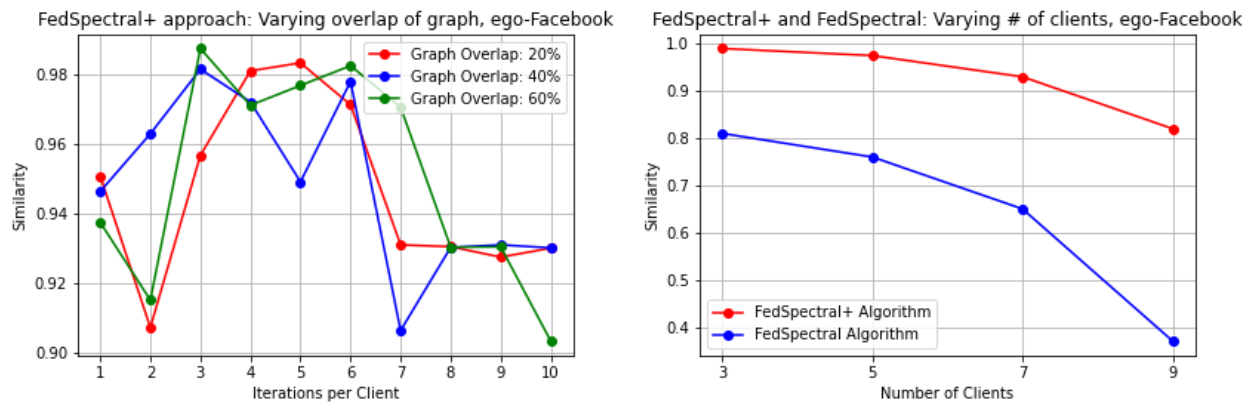}
\caption{The figure a) on the left, shows the plot of how the clustering similarity Vs iteration per client changes with respect to the overlap of the graph on \textit{ego-Facebook} dataset. The figure b) on the right, shows the plot of how the \textit{FedSpectral} and \textit{FedSpectral+} approach performs in terms of clustering similarity when the number of clients is varied on \textit{ego-Facebook} dataset.}
\label{fig:3.jpg}
\end{figure*}

In Figure \ref{fig:1.jpg} b) shows the variation in similarity with the varying number of global rounds for \textit{ego-Facebook} and \textit{email-Eu-core} dataset with constant 10 iterations per client on the \textit{FedSpectral+} algorithm with graph distributed using 40\% overlap.
\begin{itemize}
    \item For the \textit{ego-Facebook} dataset, we can observe from the plot that as the number of global rounds increases, the similarity increases. It is because the higher the number of global rounds, the higher the convergence of \textit{FedSpectral+} method.
    \item For the \textit{email-Eu-core} dataset, we can observe from the plot that as the number of global rounds increases, similarity remains constant since the algorithm converges in the first global round.
\end{itemize}

Figure \ref{fig:2.jpg} a) shows how the similarity plot changes with the number of global rounds for \textit{FedSpectral+}. We can observe that for a fixed number of iterations, as the number of global rounds increases, the similarity increases. This is because the method converges better with more global rounds.\\
Figure \ref{fig:2.jpg} b) shows how the similarity plot changes with the number of clusters for the \textit{FedSpectral+} algorithm. Since the \textit{ego-Facebook} dataset has 10 communities, we can observe that for a fixed number of iterations, the similarity is highest for 10 clusters in general.\\

Figure \ref{fig:3.jpg} a) shows how similarity plot changes with the overlap of graph data distribution for the \textit{FedSpectral+} algorithm. We can observe that with higher graph overlap, the similarity increases in general. It is because the local embedding becomes more informative because of the availability of more information about the global graph. Therefore, in the aggregation step at the global server, the deviation in the local embeddings from different clients is smaller, resulting in better similarity.\\

In Figure \ref{fig:3.jpg} b) shows the comparison of \textit{FedSpectral} approach with the \textit{FedSpectral+} method. We can see that as the number of clients increases, and if we keep the overlap of graph data within the clients to be fixed, then the similarity drops. It is because the graph structure present with a particular client becomes sparsed as the edges are distributed over more clients. We can also observe that \textit{FedSpectral+} approach works significantly better than the baseline - \textit{FedSpectral}.\\
\begin{figure*}[t]
\includegraphics[width=\linewidth, height=2.25in]{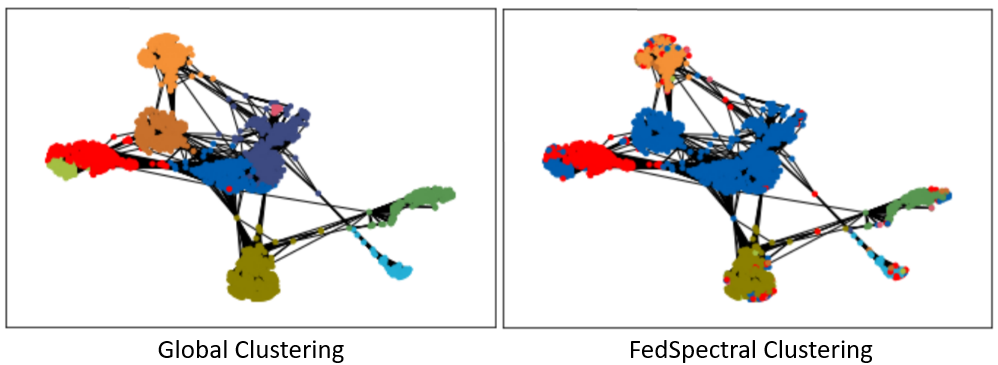}
\caption{Visual representation of \textit{ego-Facebook} dataset using \textit{FedSpectral} approach on right, and the global clustering on left.}
\label{fig:4.jpg}
\end{figure*}

\begin{figure*}[t]
\includegraphics[width=\linewidth, height=2.25in]{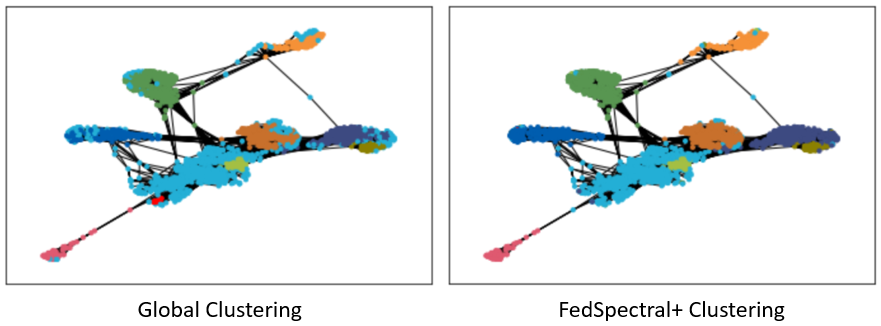}
\caption{Visual representation of \textit{ego-Facebook} dataset using \textit{FedSpectral+} approach on right, and the global clustering on left.}
\label{fig:0.jpg}
\end{figure*}
For the visual comparison, we used the \textit{ego-Facebook} dataset. We can observe that in Figure \ref{fig:4.jpg}, the \textit{FedSpectral} approach cannot distinguish all the clusters in comparison to the global clustering. It is because the \textit{FedSpectral} approach disregards that a particular node in a client may not have information about all the neighboring nodes which are spread over other clients. However, in Figure \ref{fig:0.jpg}, the \textit{FedSpectral+} approach is able to get the highest detailing in its clustering of the \textit{ego-Facebook} dataset. This is because the embedding is constructed using the power iteration method on each client and then aggregated by the server to learn the embedding of the global graph.\\

\section{Conclusion} 
To conclude, in this work, we proposed a state-of-the-art approach - \textit{FedSpectral+}, a novel algorithm for federated spectral clustering. We used the power iteration method to iteratively learn the eigenvector embedding of the global graph. As there is no exchange of raw data between clients and the server, the proposed approach overcomes the main issue of spectral clustering in the distributed setting by providing data privacy. We also give a detailed analysis of factors impacting the \textit{FedSpectral+}. As validated from the experiments, the similarity of clustering obtained using the proposed approach was comparable to that of global clustering. Thus, proving the efficacy of our approach.

\section{Future Work}
The suggested approach was only evaluated for its performance. In the future, we plan to test how resilient our algorithm is against different adversarial attacks, such as inference and poisoning attacks \cite{shen2020distributed}. We also intend to assess the possibility of a data breach while transmitting local updates to the server. To further strengthen data privacy, we will improve our method by combining it with differentially private algorithms and secured multiparty computation. 

\section{Acknowledgement}
We would like to thank Prof. Anirban Dasgupta (IIT Gandhinagar) for his continuous support and guidance throughout the research.

\bibliography{LaTeX/aaai23} 




\end{document}